\pdfoutput=1
\documentclass[11pt]{article}

\usepackage{ACL2023}
\usepackage{times}
\usepackage{latexsym}
\usepackage{xurl} 
\usepackage[T1]{fontenc}
\usepackage[utf8]{inputenc}
\usepackage{microtype}

\usepackage{tabularx}
\usepackage{array}
\newcolumntype{Y}{>{\raggedright\arraybackslash}X}
\newcolumntype{R}{>{\raggedleft\arraybackslash}X}

\usepackage{multirow}
\usepackage{inconsolata}
\usepackage{listings}
\usepackage{xcolor} % for color definitions if not already included

\usepackage{booktabs}
\usepackage{longtable}
\usepackage{rotating} 
\title{Going over Fine Web with a Fine-Tooth Comb:\\ Technical Report of Indexing Fine Web for Problematic Content Search and Retrieval }

\author{
  \textbf{Inés Altemir Marinas \thanks{ ~The main contributing author} \textsuperscript{1}},
  \textbf{Anastasiia Kucherenko\textsuperscript{2}}, 
  \textbf{Andrei Kucharavy\textsuperscript{3}}
\\
  \textsuperscript{1}École Polytechnique Fédérale de Lausanne, Switzerland\\
  \textsuperscript{2}Institute of Entrepreneurship and Management, HES-SO Valais-Wallis, Switzerland \\
  \textsuperscript{3}Institute of Informatics, HES-SO Valais-Wallis, Switzerland
}
\begin{document}
\maketitle
\begin{abstract}
Large language models (LLMs) rely heavily on web-scale datasets like Common Crawl, which provides over 80\% of training data for some modern models. However, the indiscriminate nature of web crawling raises challenges in data quality, safety, and ethics. Despite the critical importance of training data quality, prior research on harmful content has been limited to small samples due to computational constraints. This project presents a comprehensive framework for indexing and analyzing entire LLM training datasets using an ElasticSearch-based pipeline. We apply it to SwissAI’s FineWeb-2 corpus (1.5TB, four languages), achieving fast query performance—most searches in milliseconds, all under 2 seconds. Our work demonstrates the feasibility of comprehensive, real-time dataset analysis, offering practical tools for safer, more accountable AI systems.

\end{abstract}

\section{Introduction}
The capabilities of large language models \citet{brown2020languagemodelsfewshotlearners}), \citet{touvron2023llamaopenefficientfoundation}, \citet{grattafiori2024llama3herdmodels}, \citet{abdin2024phi3technicalreporthighly}, \citet{bubeck2023sparksartificialgeneralintelligence} are primarily derived from pretraining on extense web-scale datasets such as Common Crawl and derivates C4 \citet{JMLR:v21:20-074} and FineWeb \citet{penedo2024finewebdatasetsdecantingweb}. These web-scraped datasets have become the de facto standard for internet-scale pretraining datasets, as some models such as GPT-3 \citet{brown2020languagemodelsfewshotlearners} had more than 80\% of its tokens coming from Common Crawl. \\

However, the scale that enables these remarkable capabilities also introduces significant challenges. Web-mined corpora inevitably contain substantial amounts of undesirable content, including hate speech, sexually explicit material, misinformation, copyrighted data, and personally identifiable information\citet{luccioni}, \citet{mendu2025saferpretraininganalyzingfiltering}, which propagate into LLM output. Research has consistently demonstrated that such content persists even after conventional filtering procedures (such as perplexity-based filtering), with studies finding that 2-6\% of web pages in Common Crawl contain hate speech and approximately 2\% contain sexually explicit content (\citet{luccioni}). 

The implications of this contamination vary significantly across different model deployment paradigms. While API-only models can implement generation-time filtering to mitigate harmful outputs, this approach becomes ineffective for open-weights models where the full model parameters are publicly available. For open-source models, comprehensive training data mapping and analysis represents the primary defense against propagating problematic content, making preemptive dataset governance essential rather than optional. This fundamental difference underscores the critical importance of thorough training data analysis for the growing ecosystem of open-weights language models.

The personally identifiable information present in these large-scale datasets poses a severe privacy risk, as demonstrated by \citet{carlini2021extractingtrainingdatalarge}, who showed that adversaries can perform training data extraction attacks to recover individual training examples—including personally identifiable information like names, phone numbers, and email addresses—by simply querying large language models like GPT-2. These privacy concerns are further compounded by copyright and content authenticity challenges that have emerged from the indiscriminate use of web-scraped data.
The legal and ethical implications of using massive internet-scraped datasets without clear rights or consent have materialized in high-profile cases, with major content creators like the New York Times now actively blocking OpenAI’s web crawler and Common Crawl from accessing their pages using robots.txt to prevent unauthorized use of their content in AI training \citep{inproceedings}. 
Simultaneously, the infiltration of misinformation into training datasets represents another critical vulnerability. Malicious actors can deliberately inject false or misleading information into web content that subsequently gets incorporated into model training data, potentially compromising the reliability of AI systems at scale. An example of this is the Moscow-based disinformation network “Pravda”, who with 3,600,000 articles in 2024 have flooded search results and web web crawlers with pro-Kremlin discourse\footnote{\url{https://www.newsguardtech.com/special-reports/moscow-based-global-news-network-infected-western-artificial-intelligence-russian-propaganda/}}. 

These interconnected challenges—privacy violations, copyright infringement, and data contamination—underscore the urgent need for transparent, searchable, and auditable data pipelines.

A critical limitation of existing research on training data quality is the reliance on small-scale sampling due to computational constraints. \citet{luccioni} analyzed only 1\% of Common Crawl files (approximately 81 GB) due to "hardware constraints",  while \citet{mendu2025saferpretraininganalyzingfiltering} examined 1 million random web pages for their comprehensive harm taxonomy development. While these samples may be statistically representative, the limitation to subsets potentially obscures important patterns in harmful content distribution, rare but significant toxic clusters, and domain-specific bias patterns that emerge only at scale.

To address these critical challenges, this project develops a comprehensive ElasticSearch-based framework for indexing and searching complete LLM training datasets rather than representative samples. The indexing pipeline ingests large training datasets and builds searchable indexes that support multiple query paradigms: exact phrase matching for precise content identification, approximate matching through configurable fuzzy search for handling variations and typos, and semantic similarity search for discovering conceptually related content. These capabilities are unified through additive query logic, enabling complex boolean combinations that can simultaneously search for exact terms, semantic concepts, and fuzzy variants—creating an extremely powerful tool for ensuring LLM user safety and comprehensive dataset governance.

Unlike previous research constrained by computational limitations, our framework enables systematic content analysis across entire multilingual training corpora. The urgency of such comprehensive-scale analysis is underscored by the multilingual nature of modern LLM training, where harmful content patterns may vary significantly across languages and cultural contexts—variations that sampling-based approaches cannot reliably capture. While considerable research has focused on post-training alignment techniques, significantly less attention has been devoted to the foundational role of training data quality in ensuring model safety, making scalable indexing infrastructure essential for responsible AI development.

Our work specifically addresses this gap by applying our framework to SwissAI's multilingual FineWeb-2 corpus, demonstrating comprehensive exploration capabilities that sampling-based approaches cannot reliably access at scale. The contributions of this work are threefold: (1) the development of a scalable indexing infrastructure that overcomes computational barriers limiting previous research to small samples, (2) demonstration of the system's efficiency through targeted searches across multiple query types that reveal problematic content across languages in the complete training corpus, and (3) the provision of production-ready tools for ongoing dataset governance at the scale required for modern LLM development.\\
 
This framework positions Switzerland and its SwissAI model as leaders in responsible AI development, demonstrating how comprehensive training data governance can ensure AI systems are built on well-understood, ethically-sourced, and legally-compliant datasets—goals that align closely with Switzerland's longstanding values. By developing the tools and methodologies to thoroughly understand and audit training data, we can ensure that the next generation of AI systems serves society while respecting privacy, copyright, and safety principles. Crucially, our open indexing infrastructure enables independent third parties to audit both the training data and resulting models, fostering the kind of transparent, verifiable AI development that can build public trust and establish international standards for responsible AI governance.

\section{ElasticSearch}
Elasticsearch (ES) is a distributed, RESTful search and analytics engine built on Apache Lucene. 

For the distributed character of ES, we can run an ES instance in \verb|discovery.type=single-node| or 
\verb|discovery.type=multi.node|, 
where multi-node configuration allows ES to discover other nodes when forming a cluster and allows other nodes to join the cluster later.  (explain what a cluster is). In multi-node mode, the data is automatically split across multiple nodes (servers) in a cluster. Each index is divided into "shards" - independent pieces that can be placed on different machines. For now, we have explored the single-node configuration only, as the multi-node approach is pending on how feasible the true clustering across SLURM-managed nodes can be (\ref{sec:netowrk-settings}). 

\subsection{General overview of pipeline}
\begin{itemize}
\item \textbf{Text processing stage}: Implements a multi-analyzer approach that processes the 'text' field through different levels of linguistic processing - from heavily normalized text (lowercase, ASCII folding, stemming, stopword removal) to minimally processed exact matches. URL metadata is extracted and indexed for document provenance tracking. The implementation is able to handle both English and multilingual datasets

\item \textbf{Multi-field indexing strategy}: Creates three searchable versions of each text document: (1) a main analyzed field optimized for full-text search, (2) a keyword field for exact string matching, and (3) an exact field that preserves original structure while handling HTML normalization. This enables different search strategies from fuzzy matching to precise phrase detection.

\item \textbf{Inverted index construction}: Elasticsearch builds optimized inverted indexes mapping each unique term to all documents containing it, including positional information. This enables both term-based queries (finding documents with any search terms) and phrase-based queries (finding exact word sequences), with configurable relevance scoring using TF-IDF and BM25 algorithms.

\item \textbf{Distributed storage and processing}: The system uses a N-shard configuration with no replicas for optimal indexing performance. Compression and refresh interval optimizations further enhance throughput during bulk operations. Larger datasets and data directories are splitted, enabling parallel indexing processing across multiple workers. A merging operation is later executed to create a unified unique index. 

\item \textbf{Advanced query execution engine}: Supports six distinct query types ranging from simple term matching to complex boolean combinations and fuzzy matching.

\end{itemize}

\subsection{Indexing operation}
The Elasticsearch indexing component processes large-scale parquet datasets containing web-scraped text data and transforms them into searchable indexes. 
The pipeline is designed to handle massive datasets while maintaining memory efficiency and enabling distributed processing.

The indexing operation begins by streaming parquet files to prevent memory accumulation, extracting text content along with metadata (including URLs for provenance tracking). Each document undergoes multi-level text processing through configurable analyzers that create different searchable representations - from heavily normalized text optimized for semantic search to exact matches preserving original structure.

The system employs a distributed architecture with configurable sharding and supports parallel processing through file-range indexing, allowing multiple workers to process different portions of the dataset simultaneously. Performance optimizations include bulk indexing with configurable batch sizes, and dynamic refresh interval adjustments to maximize throughput during large-scale operations.This optimization is essential due to the time and memory constraints imposed on Clariden jobs. 

Advanced Elasticsearch cluster management capabilities enable merging of distributed indexes through a multi-cluster approach, where individual data directories run separate Elasticsearch instances that are then combined using remote \verb|_reindex| operations. This ensures data integrity while scaling to handle datasets that exceed single-node capacity limitations.

\subsubsection{Parameters tuning}
\label{sec:parameter-tuning}
The indexing operation is based around the \verb|elasticsearch.helpers.parallel_bulk| function, which enables concurrent bulk indexing of documents to Elasticsearch by distributing bulk requests across multiple threads

Performance optimization in runtime and memory usage requires careful tuning of multiple interdependent parameters.  

The \textbf{thread count} is the size of the threadpool to use for the bulk requests, and it should not exceed available CPU cores. Each thread count maintains its own request queue, multiplying the memory usage. In most of our experimental runs, we run with \verb|--cpus-per-task=8|. Hihger values improve the parallel processing capacity.  

The \textbf{chunk size} is the number of documents bundled in one bulk request to ES. A too small value may imply more HTTP overhead and reduce throughput. A too large value will increase memory pressure, causing potential timeouts. 

The \textbf{max chunk bytes} is the maximum size of individual bulk requests. Larger chunks will reduce HTTP overhead, but also require more RAM per thread. 

The \textbf{queue size} is  size of the task queue between the main thread (producing chunks to send) and the processing threads. (values tested have been between 2 - 8) \\

The parameter tuning analysis focuses on the FineWeb CC-MAIN-2024-51  \footnote{\url{https://huggingface.co/datasets/HuggingFaceFW/fineweb/viewer/CC-MAIN-2024-51}} dataset indexing operation.
Computing the average size of the documents, let's call it \verb|avg_doc_size|, along with \verb|thread_count|, \verb|chunk_size| and \verb|queue_size|, we can estimate this memory usage. 
For starters, we have a first constraint on \verb|chunk_size|, as this is dependent on the \verb|max_chunk_size| and the \verb|avg_doc_size|. Therefore, the division of these two values gives us an upper bound for \verb|chunk_size|.
\begin{equation}
    chunk\_size \leq \frac{max\_chunk\_size}{avg\_doc\_size}
\end{equation}
In our case, with \verb|avg_doc_size|=4KB, and setting \verb|max_chunk_size| to 50MB, we get \verb|chunk_size|=12500. 

The values of \verb|thread_count| and \verb|queue_size| can be guided by our desired memory usage.

\subsubsection{Tests}
We seek to gain insight into the scaling behavior of the indexing process. Using the FineWeb CC-MAIN-2024-51 dataset, we indexed portions of increasing size—2GB, 178GB, and the full 363GB—and observed running times of 391s, 18,000s, and 31,094s respectively.

\begin{table*}[htbp]
\centering
\footnotesize
\begin{tabularx}{\textwidth}{R R R R R R R}
\textbf{CPUs} & \textbf{Chunk size} & \textbf{Thread count} & \textbf{Queue size} & \textbf{Running time (s)} & \textbf{Peak memory (GB)} & \textbf{Index size (GB)} \\
8  & 12500  & 2  & 2  & 34.634 & 1.55  & 612.01 \\
8  & 12500  & 4  & 2  & 33.704 & 2.81  & 595.60 \\
8  & 12500  & 4  & 4  & 31.427 & 2.59  & 621.69 \\
8  & 12500  & 4  & 8  & 34.433 & 3.32  & 615.14 \\
8  & 12500  & 8  & 8  & 31.094 & 3.76  & 604.31 \\
16 & 12500  & 16 & 4  & 33.227 & 7.79  & 607.59 \\
\end{tabularx}
\caption{Performance metrics of runtime and peak memory usage for various index parameters configurations }
\label{tab:performance-metrics}
\end{table*}

Table \ref{tab:performance-metrics} summarizes the performance metrics of different indexing configurations applied to the Fineweb CC-MAIN-2024-51 dataset (original size: 363GB, available on Hugging Face). We report the running time, peak memory usage, and index size obtained when varying CPU thread count, chunk size, and queue size. These results help illustrate the trade-offs between speed, memory consumption, and index size under different parallelization settings.
We show that increasing the thread count and queue size generally reduces running time, but at the cost of significantly higher peak memory usage. Conversely, configurations with fewer threads and smaller queues yield lower memory consumption, though with longer processing times.

\subsubsection{Indexing SwissAI Filtered Fineweb 2 dataset}
\label{sec:index-fineweb2}
As the objective of this project is to develop a tool capable of analyzing and searching large-scale training datasets, and to apply such a tool in the context of the ongoing development of the 
SwissAI large language model (LLM), we present an initial indexing analysis intended as a sample case study demonstrating the tool’s potential use. \\

This analysis serves as a first step towards illustrating how the tool can support the inspection and evaluation of training data in practice. \\
The SwissAI project has produced a filtered version of the Fineweb 2 dataset as part of the pretraining data recipe. We index pertinent subdatasets of the SwissAI FineWeb 2 dataset, specifically focusing on the Italian (\verb|ita_Latn|), German (\verb|deu_Latn|), Swiss German (\verb|gsw_Latn|), and French (\verb|fra_Latn|) subsets, as these correspond to the official languages of Switzerland. The sizes of these indexed datasets are: 329 GB for \verb|ita_Latn|, 634 GB for \verb|deu_Latn|, 113 MB for \verb|gsw_Latn|, and 515 GB for \verb|fra_Latn|.

The parallel processing architecture successfully indexed 1.5TB+ of multilingual web content using file range partitioning, simultaneous batch processing, and remote reindex merging. The German dataset (634GB) demonstrated optimal performance with 16-shard configuration, achieving 79.25 GB/hour throughput and nearly 8-hour total processing time. In contrast, the French dataset (515GB) suffered from suboptimal 7-shard design, resulting in 2.3× slower processing (34.3 GB/hour) and violating Elasticsearch's 50GB/shard recommendation. Peak memory usage remained below 5.17GB across all operations, with batch sizes of 100-125 parquet files proving optimal. All details can be found in Table \ref{tab:index-metrics}.  The indexing parameters used for all these indexing operation was the optimal configuration found in \ref{sec:parameter-tuning}.

\begin{table*}[ht]
\centering
\caption{Elasticsearch Indexing Performance for FineWeb Dataset Languages}
\label{tab:elasticsearch_performance}
\resizebox{\textwidth}{!}{%
\begin{tabular}{lrrrrrrr}
\toprule
\textbf{Dataset} & \textbf{Original Size} & \textbf{Parquet Files} & \textbf{Batch} & \textbf{Peak Memory} & \textbf{Execution Time} & \textbf{Index Size} & \textbf{Merge Time} \\
 & \textbf{(GB)} & \textbf{(Total)} & \textbf{Size} & \textbf{(GB)} & \textbf{(seconds)} & \textbf{(GB)} & \textbf{(minutes)} \\
\midrule

\multirow{8}{*}{deu\_Latn} 
& \multirow{8}{*}{634} & \multirow{8}{*}{994} & 125 & 2.38 & 5587 & 95.1 & \multirow{8}{*}{379.8} \\
& & & 125 & 2.74 & 5332 & 101.2 & \\
& & & 125 & 2.84 & 5402 & 82.1 & \\
& & & 125 & 3.56 & 5142 & 85.0 & \\
& & & 125 & 3.11 & 2819 & 61.0 & \\
& & & 125 & 1.42 & 1410 & 18.6 & \\
& & & 124 & 2.22 & 1572 & 23.4 & \\
& & & 124 & 1.52 & 1298 & 17.0 & \\
\midrule

\multirow{8}{*}{fra\_Latn} 
& \multirow{8}{*}{515} & \multirow{8}{*}{785} & 100 & 4.43 & 14859 & 70.0 & \multirow{8}{*}{571.4} \\
& & & 100 & 5.17 & 20371 & 69.2 & \\
& & & 100 & 4.42 & 15014 & 69.0 & \\
& & & 98 & 4.41 & 11774 & 59.8 & \\
& & & 98 & 4.00 & 5775 & 36.2 & \\
& & & 98 & 2.61 & 2480 & 21.0 & \\
& & & 98 & 2.06 & 2152 & 17.8 & \\
& & & 93 & 1.55 & 1040 & 9.1\textsuperscript{*} & \\
\midrule

ita\_Latn & 329 & 486 & Single & 3.94 & 17148 & 213.0 & -- \\
\midrule

gsw\_Latn & 0.113 & 1 & Single & 1.92 & 25.48 & 0.35 & -- \\

\bottomrule
\end{tabular}
}
\label{tab:index-metrics}
\end{table*}

The indexing pipeline utilized optimized parameters derived from systematic tuning, including: custom text analyzers, compressed storage with \verb|best_compression codec| and disabled refresh intervals during bulk indexing.

This architecture successfully indexed over 1.5TB of multilingual web content while maintaining sub-6GB memory footprints per process, demonstrating the viability of Elasticsearch for large-scale document retrieval systems in resource-constrained environments.

\subsubsection{Upper limit of indexable information}
With the current achieved optimization, the practical upper limit for single batch-processings appears to be approximately around 500GB. (respecting Clariden's job time limit). 

Regarding ElasticSearch's shard constraints, it imposes fundamental limitations on index scalability, with a maximum number of documents per shard (~2.1 billion) and a recommended shard size: 50GB maximum, 200M documents optimal

\subsubsection{Improvements}
\begin{itemize}
    \item Dynamic sharding configuration to achieve optimal data distribution per shard based on dataset characteristics
    \item Adaptive chunk sizing in parallel batch processing operations to optimize the balance between maximum chunk size and average document size
    \item Chunk size parameter tuning to determine optimal values for maximum chunk bytes based on dataset characteristics in single-process jobs
    \item Resource monitoring for memory usage and bandwidth utilization to optimally adjust queue size and thread count parameters
\end{itemize}

Regarding production deployment enhancements, enabling the memory mapping functionality (as discussed in \ref{sec:mem-mapping}) would provide for enhanced performance. Additionally, managing parallel job submissions in SLURM through automated array job would prove useful. 

An ablation regarding the memory usage scenarios for different total memory allocation strategies

\subsection{Search operation}
The Elasticsearch search component provides a comprehensive query execution framework designed to perform diverse search operations on large-scale indexed web content. The system enables systematic exploration of massive text corpora through multiple search strategies, each optimized for different information retrieval scenarios. \\ 

In the current set-up, the search system takes predefined collections of search terms provided by a user (such as misinformation keyword lists or domain-specific vocabularies in our test case) and executes queries for each term against the complete indexed dataset. This systematic approach processes entire keyword files containing hundreds or thousands of terms, ensuring exhaustive coverage across millions of documents. The pipeline handles high-volume query execution through optimized connection management and configurable timeouts designed for large-scale index operations. \\ 
Beyond returning simple hit counts, the pipeline extracts highlighted text snippets from matching documents, capturing the actual context where search terms appear. This enables us to examine not only whether specific content exists within the corpus, but also how terms are used, what surrounding context they appear in, and how meaning varies across different documents and sources. \\ 
The system tracks query execution times, hit distributions, and relevance scoring. This monitoring reveals which query types work best for specific content domains and helps identify optimal search approaches for different research objectives. 
The relevance scoring uses BM25 (Best Matching 25) as its default algorithm, which is an improved version of TF-IDF that addresses the saturation problem where term frequency has diminishing returns and incorporates document length normalization. BM25 calculates relevance scores by combining term frequency (how often query terms appear in a document), inverse document frequency (how rare terms are across the corpus), and document length normalization. 

\subsubsection{Type of queries possible}

The index uses a \verb|web_content_analyzer| that applies:
\begin{itemize}
    \item HTML stripping: Removes HTML tags from content
    \item Standard tokenization: Splits text into tokens
    \item Lowercase normalization: Converts all text to lowercase
    \item ASCII folding: Converts accented characters (\verb|é| e)
    \item English stop word removal: Removes common words (the, and, of, etc.)
    \item English stemming: Reduces words to root forms (running run, climates climat)
\end{itemize}
This preprocessing significantly affects how queries match against indexed content.There is also a \verb|exact_match_analyzer| available, which does not perform ascii folding, stop word removal or stemming in order to identify pure verbatim. \\

The match query \verb|match_query| implements OR logic between terms, applying full-text analysis to both query and indexed content. For single-word queries like “climate,” the term is analyzed and stemmed to “climat,” which matches documents containing variations like “climate,” “climates,” or “climatic.” BM25 scoring is used for relevance ranking. For multi-word queries such as “climate change,” each word is analyzed separately, resulting in a query for “climat” OR “chang.” Documents need to contain only one of the terms to match, and word order is irrelevant; “change climate” yields the same results. Documents containing both terms receive higher relevance scores.

The match phrase query \verb|match_phrase_query| requires exact phrase matching, with configurable proximity tolerance. For single-word queries, it behaves like the match query. For multi-word queries like “climate change,” the terms must appear in the exact order after analysis. For example, “climat” must immediately precede “chang” when the slop is 0, preserving phrase integrity. The \verb|MATCH_PHRASE_SLOP| parameter controls tolerance for word proximity; slop 0 enforces exact matching, slop 1 allows one intervening word, and slop 2 allows two. This means slop 1 can match “climate and change,” while slop 2 can match “climate action and change.”

The term query exact \verb|term_query_exact| bypasses text analysis entirely, searching for exact tokens using the text.exact field. For single-word queries, no stemming or analysis is applied, so searching for “climate” only finds the exact token “climate,” excluding variations like “climates” or “climatic.” This is useful for exact word forms, proper nouns, or technical terms. Term queries do not support multi-word inputs and will return empty results for such cases, as they operate on single tokens only.

The fuzzy query  \verb|fuzzy_query| handles typographical errors and variations through edit distance calculations. For single-word queries, it uses Levenshtein distance to find similar terms within a tolerance, typically allowing one or two character edits. For example, searching for “climat” could match “climate” or “climax” if within tolerance. Text analysis still applies. For multi-word queries, the system falls back to a multi-match query with fuzziness enabled per term, using AND or OR operators. This approach is more computationally expensive and can match inputs like “changge” with “change” or “cliamte” with “climate.”

The boolean must query \verb|bool_must_query| creates structured boolean queries with configurable logic and scoring control. For single-word queries, it creates a single must clause, similar to a match query with the AND operator but allowing complex scoring. For multi-word queries, it creates separate match clauses where all terms must be satisfied, up to a configurable maximum (default is three terms). The \verb|BOOL_MUST_OPERATOR| parameter controls term relationships, where “and” requires all terms and “or” allows partial matches with \verb|minimum_should_match| . This provides fine-tuned control over precision and recall.

\subsubsection{Verbatim query length search time}
\begin{table}[htbp]
\centering
\footnotesize
\renewcommand{\arraystretch}{2} % Increase row height by 20%
\begin{tabularx}{\linewidth}{>{\raggedright\arraybackslash}p{0.9cm} c c c c}
\textbf{Query Length} & \textbf{Avg (ms)} & \textbf{Med (ms)} & \textbf{Std (ms)} & \textbf{Hit Rate (\%)} \\
\midrule
1 word    & 36.07 & 27.87 & 41.09 & 98.7 \\
10 words  & 15.01 & 7.30  & 14.73 & 100.0 \\
100 words & 55.51 & 40.86 & 35.64 & 100.0 \\
300 words & 100.51 & 105.50 & 20.52 & 99.5 \\
\end{tabularx}
\renewcommand{\arraystretch}{1} % Reset to default
\caption{Query performance statistics across different segment lengths for the GSW index. Results show average (Avg), median (Med), and standard deviation (Std) of query execution times, plus hit rates indicating successful query matches.}
\label{tab:query_performance_detailed}
\end{table}

The purpose of this experiment is to evaluate the performance of match phrase queries when applied to an index with varying query segment length. Specifically, this experiment seeks to explore the feasibility of using verbatim phrase matching as a method for detecting potential memorization and regurgitation of training data passages by the model.

To this end, we sampled random contiguous segments of varying lengths (1, 10, 100, and 300 words) from the source datasets used to build the indexes. These segments were then issued as match phrase queries against the corresponding indexes. 

Table~\ref{tab:query_performance_detailed} summarizes the query performance results. For the GSW index, the hit rate remains consistently high across all segment lengths, ranging from 98.7\% for 1-word queries to 100\% for 10- and 100-word queries, and 99.5\% for 300-word queries. Query execution times exhibit the expected trend of increasing with segment length: median query times range from approximately 7 ms for 10-word queries to 105 ms for 300-word queries. The standard deviation of execution time also generally increases with query length, though the 300-word queries display a relatively stable latency (standard deviation of 20.52 ms), suggesting that longer phrases may incur a more predictable computational cost within this small index.

Overall, these preliminary results indicate that match phrase queries can be executed with low latency on smaller indexes even for longer segments, supporting the feasibility of this method as part of a toolset for detecting potential memorization. Further experimentation on larger indexes and additional languages will be required to fully characterize performance and accuracy under conditions representative of production-scale language model training datasets.

\section{Search results on SwissAI training data}
We present an initial analysis intended as a sample case study demonstrating the tool’s potential use. This analysis serves as a first step towards illustrating how the tool can support the inspection and evaluation of training data in practice.

\subsection{SwissAI FineWeb-2 Dataset}
On the German, French, Swiss German and Italian datsets from FineWeb-2, indexed in \ref{sec:index-fineweb2}, we conduct searches using the Weaponized Words dictionary \footnote{\url{https://weaponizedword.org/}}, which provides language-specific lexicons of curse words, slurs, and other forms of toxic or discriminatory language. The motivation behind this analysis is to explore the presence and distribution of potentially harmful or offensive terms within the SwissAI filtered training data, and to validate our experimental pipeline. An example of the result of queries of the german lexicon of WeaponizedWords on the DEU index is Table \ref{tab:query_hits_clean}. \\

\begin{table*}[!ht]
\centering
\begin{tabular}{l|lrrrr}
\hline
\textbf{Index (Size)} & \textbf{Query Type} & \textbf{Avg Time (ms)} & \textbf{Med Time (ms)} & \textbf{Avg Hits} & \textbf{Total Queries} \\
\hline
\multirow{5}{*}{\textbf{GSW (389MB)}} 
  & match\_query          & 10.75 & 7.61 & 210.02 & 53 \\
  & match\_phrase\_query  & 10.71 & 7.84 & 21.34 & 53 \\
  & term\_query\_exact    & 5.74  & 4.75 & 14.68 & 53 \\
  & fuzzy\_query          & 38.45 & 35.00 & 520.35 & 52 \\
  & bool\_query           & 9.46  & 7.46 & 210.02 & 53 \\
\hline
\multirow{5}{*}{\textbf{DEU (400GB)}} 
  & match\_query          & 185.49 & 136.95 & 7754.94 & 53 \\
  & match\_phrase\_query  & 86.62  & 51.56  & 7590.21 & 53 \\
  & term\_query\_exact    & 177.06 & 198.71 & 5290.13 & 53 \\
  & fuzzy\_query          & 3462.92 & 3586.34 & 8636.56 & 52 \\
  & bool\_query           & 76.97  & 51.94  & 7754.94 & 53 \\
\hline
\multirow{5}{*}{\textbf{FRA (313GB)}} 
  & match\_query          & 191.45 & 120.03 & 9162.91 & 75 \\
  & match\_phrase\_query  & 193.11 & 56.44  & 7607.07 & 75 \\
  & term\_query\_exact    & 106.85 & 135.62 & 4479.76 & 75 \\
  & fuzzy\_query          & 2055.44 & 2333.66 & 9382.21 & 57 \\
  & bool\_query           & 71.87  & 53.14  & 9162.91 & 75 \\
\hline
\multirow{5}{*}{\textbf{ITA (196GB)}} 
  & match\_query          & 156.42 & 100.78 & 7359.83 & 24 \\
  & match\_phrase\_query  & 70.93  & 31.80  & 5887.25 & 24 \\
  & term\_query\_exact    & 108.91 & 112.44 & 5276.42 & 24 \\
  & fuzzy\_query          & 1175.49 & 1187.93 & 8754.60 & 20 \\
  & bool\_query           & 41.80  & 32.01  & 7359.83 & 24 \\
\hline
\end{tabular}
\caption{Query performance comparison across language indexes (and their index size). Metrics include average and median query times (ms), average hits per query, and total queries executed.  Queries done in the respective language WeaponizedWords dataset. }
\label{tab:query-performance}
\end{table*}

Here is the query configuration used for all 4 indexes to perform the queries on WeaponizedWords:

\begin{lstlisting}
   
{
    "execute_match_query": true,
    "execute_match_phrase_query": true,
    "execute_term_query_exact": true,
    "execute_wildcard_query": false,
    "execute_fuzzy_query": true,
    "execute_bool_must_query": true,
    "match_query_operator": [
        "or"
    ],
    "match_phrase_slop": [
        0
    ],
    "bool_must_operator": "or",
    "bool_must_max_words": 3,
    "bool_must_minimum_should_match": "50%"
}
\end{lstlisting}

\begin{table*}[!ht]
\centering
\small
\begin{tabular}{p{2.5cm} p{1.5cm} p{12cm}}
\toprule
\textbf{Query Type} & \textbf{Score} & \textbf{Snippet (with URL)} \\
\midrule
match\_query &
21.066 &
Halt Die Fresse "Halt die Fresse!" Dann halt die Fresse. Was machen die da. Halt Die Fresse nicht buchen. - Halt die fresse! Namensräume Artikel Diskussion. (\url{https://neocarla.com/aktuelle-nachrichten-welt/halt-die-fresse.php}) \\
&
20.764 &
Daraus folgt: Halt die Fresse! Antwort: Halt die Fresse! Halt die Fresse! - Stimmt tatsächlich, echt schönes Wetter heute! Schönen Tag noch! Antwort: Halt die Fresse! Halt die Fresse! Halt die Fresse! - Ja, ich weiß. Sie hat mir heute davon erzählt, aber sie war davon nicht so begeistert. (\url{http://kamelopedia.net/wiki/Halt_die_Fresse}) \\
&
20.744 &
Das war wie ein Schlag in die Fresse. Nur halt ins Herz. wie ein schlag in die fresse, nur ins herz -.- Hauptschule: "Halt die Fresse!" Realschule: "Halt die Fresse!" Gymnasium Wäre dein Herz geborchen, wärst du Tot. Also halt die Fresse! Junge: Ach H Starrt sie auf dein Mund dann küss sie.. HALT DIE FRESSE ICH HÖR AUF Irgendwann schlag ich Dir in die Fresse. Versprochen! ;D (\url{https://www.spruchmonster.de/das-war-wie-ein-schlag-die-fresse-nur-halt-ins-herz}) \\
&
20.568 &
Bewertung, Halt die Die Geschichte Rapper Toony, dessen Beitrag für Halt die Fresse man auch diese Informationen hilfreich. Halt die Fresse man. Halt Die Fresse Man Translations \& Examples VideoOidorno - Halt die Fresse ich will saufen (Official Video) Halt Die Fresse Man Inhaltsverzeichnis VideoHDF Halt Die Fresse Man "Halt die Fresse!" (\url{https://ebony-anal.com/syrien-nachrichten/halt-die-fresse-man.php}) \\
&
20.368 &
HDF, Halt die FresseHalt deine Fresse. Mir kamen Tränen in die Augen und er meinte nur Halt die Fresse und 3. Juni 2015. Einen Beutel mit dem Aufdruck Halt die Fresse verschicken Sie mit Sticker und Konfetti. (\url{http://arrestmorning.live/halt-die-fresse-ich-liebe-dich/}) \\
\midrule
match\_phrase\_query &
20.990 &
Übersetzung im Kontext von "halt die Fresse" in Deutsch-Englisch von Reverso Context: Halt die Fresse, Jackie Boy. Halt Die Fresse "Halt die Fresse!" Als Halt Die Fresse - NavigationsmenüAmazon Web Services Cloud Computing Dienste von Amazon. Halt Die Fresse nicht buchen. - Halt die fresse! (\url{https://neocarla.com/aktuelle-nachrichten-welt/halt-die-fresse.php}) \\
&
20.741 &
Daraus folgt: Halt die Fresse! Antwort: Halt die Fresse! Halt die Fresse! - Stimmt tatsächlich, echt schönes Wetter heute! Schönen Tag noch! Antwort: Halt die Fresse! Halt die Fresse! Halt die Fresse! - Ja, ich weiß. Sie hat mir heute davon erzählt, aber sie war davon nicht so begeistert. (\url{http://kamelopedia.net/wiki/Halt_die_Fresse}) \\
&
20.344 &
Übersetzung im Kontext von "halt die Fresse" in Deutsch-Englisch von Reverso Context: Halt die Fresse, Jackie Boy. Halt Die Fresse Man Translations \& Examples VideoOidorno - Halt die Fresse ich will saufen (Official Video) Halt Die Fresse Man Inhaltsverzeichnis VideoHDF Halt Die Fresse Man "Halt die Fresse!" (\url{https://ebony-anal.com/syrien-nachrichten/halt-die-fresse-man.php}) \\
&
20.280 &
Komm halt die Fresse FRÜHER warst du ein guter Freund. sein; aber nein es heißt ja Hauptschule: "Halt die Fresse!" Realschule: "Halt die Fresse!" (\url{https://www.spruchmonster.de/und-du-sollst-ein-guter-freund-sein-und-immer-fuer-mich-da-gewessen-seinkomm-halt-die-fresse}) \\
&
20.016 &
Tasse Ich bin der Boss - Halt die Fresse! Product.Nr. cup\_kn\_ibdb product description Tasse Knorkator "Ich bin der Boss - Halt die Fresse!" die Fresse! (\url{https://www.fantotal.de/en/knorkator-shop/jewelry/tasse-knorkator-ich-bin-der-boss-halt-die-fresse.html}) \\
\bottomrule
\end{tabular}
\caption{Example of output of DEU index with 'halt die fresse' query from WeaponizedWords. Top 5 hits per query type (cleaned snippets, URL at end of snippet).}
\label{tab:query_hits_clean}
\end{table*}

\subsection{SwissAI FineWeb-edu-score 2 dataset}
We index all of the SwissAI Fineweb-edu-score 2 dataset. The dataset has undergone some filtering process, and is composed of 95 MAIN-CC dumps. As illustrated with the previous SwissAI FineWeb-2 dataset, we perform parallel processing of the 95 dumps simultaenously, which render 95 source indexes. For these indexes, at the time of writing of the raport, subsequent merge stages to obtain a unique final index are still ongoing. We randomly sample 3 source indexes among these 95 to perform search queries upon WeaponizedWords (visible in Table \ref{tab:transtrender_hits}), a list of Obscene Words \footnote{\url{https://github.com/LDNOOBW/List-of-Dirty-Naughty-Obscene-and-Otherwise-Bad-Words}}, and a Chat-GPT generated list of misinformation keywords (see Table \ref{tab:vaccines_autism_results})

\begin{table*}[!ht]
\centering
\small
\begin{tabular}{p{2.5cm} p{1.5cm} p{12cm}}
\toprule
\textbf{Query Type} & \textbf{Score} & \textbf{Snippet (with URL)} \\
\midrule
match\_query &
25.438 &
No, there's clear evidence that the MMR (measles, mumps, rubella) vaccine does not cause autism. | vaccines (but not MMR), caused autism. | More recently, some parents have been concerned that too many vaccines given too soon might cause autism. (\url{https://www.babycenter.com/404_does-the-mmr-vaccine-put-my-child-at-greater-risk-for-autism_11518.bc}) \\
&
25.437 &
Many proponents of the idea that vaccines cause autism have a habit of shifting between variants of the autism / vaccine theory without a lot of thought | It seems that they don’t care about exactly how vaccines cause autism, as long as they do. | that is caused by vaccines. (\url{http://www.aconversationonautism.com/Vaccines-and-Autism/Vaccines-Autism-Generally/Logical-Problems-with-the-Arguments/Theory-Shifting}) \\
&
25.306 &
cause autism. | Brian Hooker PhD, linking vaccines as a plausible cause of autism. | in the cause of autism Dr. (\url{http://vaccinationdecisions.net/vaccines-and-autism/}) \\
&
25.301 &
The belief that vaccines cause autism is not just wrong, but damaging. | Also, the component in vaccines that is "causing autism," hasn't been in vaccines since the 90s. | Furthermore, if you believe that vaccines cause autism, the only way a vaccine could cause autism is if the disease that the child is being vaccinated (\url{http://www.debate.org/opinions/are-vaccines-causing-autism}) \\
\midrule
match\_phrase\_query &
23.401 &
Japanese Data Show Vaccines Cause Autism June 3, 2009 Just months following the US Court of Federal Claims rejection of the claim that the MMR vaccine causes autism, here you will see data from formal peer refereed medical papers showing that vaccines caused autism in Japanese children and will be doing (\url{http://recoveringnicholas.com/2010/02/13/japanese-data-show-vaccines-cause-autism/}) \\
&
23.196 &
Does Thimerosal used in vaccines cause Autism? 2. Does the MMR vaccine cause autism? 3. | Does the combination of Thimerosal-containing vaccines and the MMR vaccine cause autism? (\url{http://spiriteddoula.blogspot.com/2009/02/vaccine-court-to-release-autism-ruling.html}) \\
&
22.865 &
May 13, 2008 WASHINGTON (AP) — Parents claiming that childhood vaccines cause autism should not be rewarded by the courts when the scientific community | Court of Claims alleging that vaccines caused autism and other neurological problems in their children. | Lawyers for the families are presenting three different theories of how vaccines caused autism. (\url{http://www.infowars.com/families-make-case-for-vaccine-link-to-autism/}) \\
&
22.754 &
Do Vaccines Cause Autism? Our crack team of scienticians have investigated. Read on for the full report. Q: Do vaccines cause autism? A: No. (\url{http://blog.westandfirm.org/2008/06/change-of-pace.html}) \\
&
22.719 &
The special "vaccine court" ruled yesterday, in three separate cases, each of which was testing a different strategy for claiming that vaccines cause autism | , that there was no evidence that vaccines caused autism in any of the cases. | The judges rule against the notion that MMR vaccine causes autism, and against the notion that thimerosal used as a preservative in vaccines causes autism (\url{http://genome.fieldofscience.com/2009/02/special-court-rules-that-vaccines-are.html}) \\
\bottomrule
\end{tabular}
\caption{Example of output of FineWeb-edu-score-2 index with "vaccines cause autism" query from misinformation dataset. Top 5 hits per query type (cleaned snippets, URL at end of snippet).}
\label{tab:vaccines_autism_results}
\end{table*}

\begin{table*}[!ht]
\vspace{-60pt}
\hspace{-2cm}
\footnotesize
\renewcommand{\arraystretch}{0.8}
\begin{tabular}{p{1.4cm} p{1cm} p{16cm}}
\toprule
\textbf{Query Type} & \textbf{Score} & \textbf{Snippet (with URL)} \\
\midrule
match &
28.392 &
What is a transtrender? Binary trans people expose "transtrenders" on Youtube. Transtrending is gatekeeping (\url{https://www.beyounetwork.org/articles/transtrending-what-it-means-and-how-it-harms}) \\
&
27.149 &
Transtrenders never experience dysphoria or oppression, often identify as "otherkin." They are gender-conforming cisgender people pretending to be transgender (\url{https://www.urbandictionary.com/author.php?author=transapple}) \\
&
26.210 &
Body dysphoria separates true transsexuals from "transtrenders." In the truscum worldview, true transsexuals deserve priority over transtrenders (\url{https://transblog.grieve-smith.com/2015/01/}) \\
&
25.143 &
Transtrend aims to contribute to well-functioning markets. Transtrend promotes balance between supply and demand (\url{https://www.transtrend.com/en/approach/responsible-investing/}) \\
&
25.085 &
A "transtrender" refers to a person who identifies as transgender because they think it's cool to do so. There's a lot of problematic implications that go with the term "transtrender." "Transtrender" is a word no person in this community should ever use or condone. (\url{https://letsqueerthingsup.com/2015/03/28/why-the-trans-community-needs-to-ban-the-word-transtrender-for-good/?like_comment=2496&_wpnonce=108d059f06}) \\
\midrule
match\_phrase &
28.392 &
What is a transtrender? Among (binary) trans people on Youtube, you can find reaction videos exposing "transtrenders" or video opinions on "transtrending". What binary trans people need to understand about transtrending Binary trans people need to understand that transtrending is gatekeeping (\url{https://www.beyounetwork.org/articles/transtrending-what-it-means-and-how-it-harms}) \\
&
27.149 &
Transtrenders never experience disphoria transphobia or opression, and often identify as "otherkin." Transtrenders are basically gender-onforming cisgender people pretending to be transgender and using different pronouns because they think it makes them (\url{https://www.urbandictionary.com/author.php?author=transapple}) \\
&
26.210 &
the word is adopted as a badge of pride by many people who espouse it), the feeling of body dysphoria separates the true transsexuals from the wannabe "transtrenders" truscum worldview, resources available for trans people are scarce, and the true transsexuals with their medical condition deserve priority over the transtrenders (\url{https://transblog.grieve-smith.com/2015/01/}) \\
&
25.143 &
Transtrend aims to contribute to well-functioning, well-organized and reliable markets. Transtrend – and its clients – must therefore be prepared to take risks. Transtrend aims to promote a healthy balance between supply and demand. (\url{https://www.transtrend.com/en/approach/responsible-investing/}) \\
&
25.085 &
A "transtrender" refers to a person who identifies as transgender because they think it's cool to do so. There's a lot of problematic implications that go with the term "transtrender." "Transtrender" is a word no person in this community should ever use or condone. (\url{https://letsqueerthingsup.com/2015/03/28/why-the-trans-community-needs-to-ban-the-word-transtrender-for-good/?like_comment=2496&_wpnonce=108d059f06}) \\

\midrule
fuzzy &
22.905 &
Trans people who long for the days where transition and dysphoria were treated seriously have coined these new invaders as "transtrenders," with the implication These new additions seem to only cause confusion amongst the general public who historically have viewed transgenderism as indicative of a gender transition (\url{https://thepostmillennial.com/transtrender-craze-trans-people}) \\
&
21.260 &
Transtrend aims to contribute to well-functioning, well-organized and reliable markets. Transtrend – and its clients – must therefore be prepared to take risks. Transtrend aims to promote a healthy balance between supply and demand. (\url{https://www.transtrend.com/en/approach/responsible-investing/}) \\
&
20.861 &
transgender issues extensively, also looks at mental health issues in the bisexual and transger communities, and contains information on bisexuality and transgernderism and particularly the common ground between the bisexual and transgender communities (\url{http://www.pinktherapy.com/en-gb/knowledge/bisexual.aspx}) \\
&
20.717 &
transgender issues extensively, also looks at mental health issues in the bisexual and transger communities, and contains information on bisexuality and transgernderism and particularly the common ground between the bisexual and transgender communities (\url{http://www.pinktherapy.com/en-gb/knowledge/bisexual.aspx}) \\
&
20.139 &
That being said, pretending to be a female is unlikely a sexual strategy due to a very low prevalence rate of transgenderism of 0,001\% Many transgenders appear to use the protected class status of transgenderism as a means of status ascension (transtrender) (\url{https://incels.wiki/w/Sneaker_male}) \\
\midrule
bool\_must &
56.784 &
What is a transtrender? Among (binary) trans people on Youtube, you can find reaction videos exposing "transtrenders" or video opinions on "transtrending". What binary trans people need to understand about transtrending Binary trans people need to understand that transtrending is gatekeeping (\url{https://www.beyounetwork.org/articles/transtrending-what-it-means-and-how-it-harms}) \\
&
54.298 &
Transtrenders never experience disphoria transphobia or opression, and often identify as "otherkin." Transtrenders are basically gender-onforming cisgender people pretending to be transgender and using different pronouns because they think it makes them (\url{https://www.urbandictionary.com/author.php?author=transapple}) \\
&
52.420 &
the word is adopted as a badge of pride by many people who espouse it), the feeling of body dysphoria separates the true transsexuals from the wannabe "transtrenders" truscum worldview, resources available for trans people are scarce, and the true transsexuals with their medical condition deserve priority over the transtrenders (\url{https://transblog.grieve-smith.com/2015/01/}) \\
&
50.286 &
Transtrend aims to contribute to well-functioning, well-organized and reliable markets. Transtrend – and its clients – must therefore be prepared to take risks. Transtrend aims to promote a healthy balance between supply and demand. (\url{https://www.transtrend.com/en/approach/responsible-investing/}) \\
&
50.170 &
A "transtrender" refers to a person who identifies as transgender because they think it's cool to do so. There's a lot of problematic implications that go with the term "transtrender." "Transtrender" is a word no person in this community should ever use or condone. (\url{https://letsqueerthingsup.com/2015/03/28/why-the-trans-community-needs-to-ban-the-word-transtrender-for-good/?like_comment=2496&_wpnonce=108d059f06}) \\
\bottomrule
\end{tabular}
\caption{Example of output of FineWeb-edu-score-2 index with 'transtrenders' query from WeaponizedWords. Top 5 hits per query type (cleaned snippets, URL at end of snippet)}
\label{tab:transtrender_hits}
\end{table*}

Further more extensive will results will be available on the HEVS Github \verb|elasticsearch_alps| repository.

\section{Deployment on ALPS Clariden - Technicalities}

This project has been deployed on Clariden, a vcluster within the Alps system of the CSCS, the Swiss National Supercomputing Centre \footnote{\url{https://www.cscs.ch/computers/alps}}. The deployment process has revealed several technical challenges stemming from the platform's specific container management approach and security constraints.

\subsection{Container environment}
Clariden provides its own container environment through the Container Engine (CE), which is specifically designed for HPC workloads and does not support Docker directly. Docker is incompatible with Clariden due to two primary constraints: security concerns arising from Docker's default shared read/write permissions for all users (rather than inheriting permissions from the user who initiated the container), and architectural incompatibility, as most Docker containers are built for x86/x86\_64/amd64 architectures rather than the arm64 architecture used by Clariden.

\subsubsection{Docker incomptability}
Since Clariden does not support the direct use of Docker, we cannot directly pull and use the official Elasticsearch image from Docker Hub\footnote{\url{https://hub.docker.com/_/elasticsearch}}. This also precludes the use of Docker Compose, which would have provided significant benefits for orchestration by allowing us to compose existing elements with ease—for example, combining an existing Elasticsearch container with storage blocks and Python code in separate containers. Docker Compose facilitates problem isolation to individual containers and simplifies component updates and upgrades by allowing modular changes to specific parts of the deployment.
This limitation means we cannot directly use the official Docker Hub image for Elasticsearch\footnote{\url{https://hub.docker.com/_/elasticsearch}}, which introduces additional complexity in building, customizing, and deploying the container. The process requires creating custom OCI-compliant container images using Podman and Containerfile, ensuring that components integrate correctly with Clariden's resource management and security policies. Furthermore, efficient resource utilization is critical—workloads must be registered within SLURM job windows, and since Alps is an extremely efficient computing platform designed for high-performance computing, components must be compiled against optimized libraries before being run at scale to maintain the platform's performance standards.

\subsubsection{Custom container image construction}
\label{sec:mem-mapping}

Following the CSCS Documentation, \footnote{\url{https://confluence.cscs.ch/spaces/KB/pages/868834153/Building+container+images+on+Alps}}, we must build custom OCI-compliant container images using Podman and Containerfile. This one will include a certain version of Elasticsearch, the corresponding Elasticsearch API, and the necessary extra tools like curl, pyarrow, python3 and so on.

Following, the Environment Definition File (EDF), a text file in the TOML format, represents the creation of a computing environment based on such container image. Environment variables defined in TOML files are not properly interpreted by Elasticsearch during startup, requiring runtime configuration through command-line parameters or configuration files.
This limitation necessitates passing critical Elasticsearch settings as command-line arguments during container startup, although this can be partially mitigated by using SLURM files tht register such arguments

A major system-level constraint is the \textbf{memory mapping limitation}. Elasticsearch relies on memory mapping (mmap) for efficient index access, requiring the kernel parameter \verb|vm.max_map_count| to be set to at least 262,144.  However, Clariden's default value is 65,530, and unprivileged containers cannot modify kernel parameters.  Without proper configuration, Elasticsearch fails during startup with:
\begin{lstlisting}
    bootstrap check failure [1] of [1]: max virtual memory areas vm.max_map_count [65530] is too low, increase to at least [262144]
\end{lstlisting}
We resolve this by disabling memory mapping entirely using:
\begin{lstlisting}
    node.store.allow_mmap=false
\end{lstlisting}
This solution enables Elasticsearch to run but introduces performance implications such as reduced I/O performance, as without memory mapping, disk I/O operations become less efficient, potentially increasing query response times and indexing throughput limitations. There is also an increased memory usage, as alternative memory management strategies consume more RAM and scalability considerations, as these performance impacts may become more pronounced as data volume and concurrent operations increase.\\

Containers launched on Clariden inherit environment variables from the host system, including \verb|JAVA_HOME|, which may be set by system configurations as pointing to cluster-managed Java installations. This creates conflicts because Elasticsearch bundles its own optimized OpenJDK distribution and the bundled JVM is specifically tuned for Elasticsearch's performance characteristics, and the host Java installations may be incompatible with containerized Elasticsearch requirements. 

The conflict manifests as:
\begin{lstlisting}
    warning: ignoring JAVA_HOME=/usr/lib64/jvm/java-11-openjdk-11; using bundled JDK
    could not find java in JAVA_HOME at /usr/lib64/jvm/java-11-openjdk-11/bin/java
\end{lstlisting}
We resolve this by explicitly managing the Java environment within the container:
\begin{lstlisting}
    unset JAVA_HOME
    export ES_JAVA_HOME="/usr/share/elasticsearch/jdk"
\end{lstlisting}
This ensures Elasticsearch uses its bundled, tested JDK while avoiding conflicts with host system Java configurations.

\subsection{Networking settings}
\label{sec:netowrk-settings}
When implementing Elasticsearch within containerized environments on the CSCS Alps Clariden cluster, to execute indexing and search operations , we may encounter network binding configuration issues, such as the
\begin{lstlisting}
Connection error: HEAD http://localhost:9200/ [status:400]
\end{lstlisting}
which bring forward the necessity to configure specific host and port variables.\\

Firstly, Elasticsearch requires specific network configuration to operate properly. In a basic setup, Elasticsearch uses two types of network communication: the HTTP interface (default port 9200) for client requests like searches and data ingestion, and the transport interface (default port 9300-9400) for internal node-to-node communication. Each node has two different network interfaces - clients send requests to Elasticsearch's REST APIs using its HTTP interface, while the transport interface is used for inter-node communication in multi-node clusters and for internal component coordination and specialized client connections even in single-node setups. By default, Elasticsearch binds only to `localhost` (127.0.0.1), which means it cannot be accessed remotely - this is a security feature. \\

This HPC architecture creates a critical incompatibility with Elasticsearch's default networking behavior, manifested as
\begin{lstlisting}
Connection error: HEAD http://localhost:9200/ [status:400]
\end{lstlisting}
The error occurs perhaps because Alps compute nodes route HTTP traffic through proxy.cscs.ch:8080. The proxy may intercept localhost requests and add headers incompatible with Elasticsearch's REST API, despite localhost being in the noproxy list. This issue can happen for a variety of different reasons (container networking complications, application level proxy settings, namespace ambiguity), but further investigation is required to clarify exactly the cause of this misbehavior, which is why this point is being clarified with CSCS engineers as of the moment of the report writing.

The solution requires explicit network configuration that bypasses the proxy and forces localhost-only binding: first, configure proxy bypass with:
\begin{lstlisting}
    export no_proxy="${no_proxy},127.0.0.1,localhost" 
    export http_proxy=""` 
    and use `curl --noproxy "127.0.0.1"` for all local API calls 
\end{lstlisting}

second, explicitly bind all Elasticsearch network interfaces to localhost using:
\begin{lstlisting}
    -E network.host=127.0.0.1 
    -E http.host=127.0.0.1 
    -E transport.host=127.0.0.1 
    -E network.bind_host=127.0.0.1 
    -E network.publish_host=127.0.0.1
\end{lstlisting}

and third, disable cluster discovery with:
\begin{lstlisting}
     -E discovery.type=single-node 
\end{lstlisting}
To comply with SLURM's job isolation model. This configuration is architecturally mandated by the intersection of Alps' proxy-mediated networking and SLURM's resource management, making it the only viable approach for successful Elasticsearch deployment in this specialized HPC environment.

\subsection{Future directions in technical deployment}

The current Elasticsearch implementation is limited to single-node clusters (discovery.type=single-node) which prevents horizontal scaling across SLURM's multi-node allocations and significantly constrains indexing performance. To optimize for SLURM workloads, the cluster setup should leverage SLURM environment variables like \verb|SLURM_NODELIST| and \verb|SLURM_JOB_NODELIST| to automatically configure proper cluster discovery with seed hosts, enabling true distributed indexing. Additionally, the numerous scattered Elasticsearch configuration parameters throughout the scripts should be consolidated into a centralized environment variable management system that can dynamically tune heap sizes, thread pools, and buffer settings based on SLURM resource allocations (\verb|SLURM_MEM_PER_NODE|, \verb|SLURM_CPUS_PER_TASK|). This would transform the current single-node bottleneck into a scalable, SLURM-native distributed Elasticsearch cluster capable of parallel indexing across multiple compute nodes.

\section{Comparison to other methods}

\textbf{Bloom filters} may offer a significantly improvement in indexing speed compared to ES, as among other factors they do not build any complex structures, just bit arrays. 
However, they present severe limitations in their search capabilities: they can only perform membership testing, quickly resolving if a term definitely doesn't exist in a dataset (no false negatives) or might exist (possible false positives). They are therefore extremely memory-efficient for large text corpora but are only useful when one needs a fast "does this term appear anywhere?" check. 
They have no storage of the document IDs, the term positions, frequencies, or actual content. Consequently, they cannot resolve questions such as: which documents contain the term?, where in documents do these terms appear?, how many times do these terms occur (ranking and frequency information), are there any term relationships or proximity? (phrase matching) and so on.

\textbf{Infinigram} is a Burrow-Wheeler Transform-based method, based on the order permutation to create an implicit pointer-based index. It generally enablew fast substring and n-gram matching across massive text corpora using pure lexical pattern matching, without any semantic understanding or meaning-based search capabilities. It operates solely on exact character sequences and n-gram patterns without supporting boolean operators like AND, OR, or NOT. In comparison, ElasticSearch has the capability to implement semantic-search and complex boolean queries that allow for a wider variety of search operations.

%\section*{Acknowledgements}
%I want to thank very dearly Dr. Kucherenko, my supervisor, and professors Pr. Kucharavy and Pr. Bosselut for their invaluable insights, guidance and time. 

\bibliography{anthology,custom}
\bibliographystyle{acl_natbib}

\appendix

%\section{Appendix}
%A step-by-step example of how to deploy ES in Clariden is available in the mentioned  HEVS Github \verb|elasticsearch_alps| repository.

\end{document}